\documentclass[a4paper]{svproc}
\usepackage{url}
\usepackage{graphicx}
\usepackage{amsmath,amssymb,amsfonts}%
\usepackage{multirow}%
\usepackage{mathrsfs}%
\usepackage[title]{appendix}%
\usepackage{xcolor}%
\usepackage{textcomp}%
\usepackage{manyfoot}%
\usepackage{booktabs}%
\usepackage{algorithm}%
\usepackage{algpseudocode}%
\usepackage{listings}%
\usepackage{hyperref}%
\usepackage{microtype}
\newcommand{\PROBLEMFORMULATI}[1]{\textcolor{black}{#1}}
\newcommand{\dope}[1]{\textcolor{black}{#1}}
\newcommand{\cvpf}[1]{\textcolor{black}{#1}}
\newcommand{\capture}[1]{\textcolor{black}{#1}}
\newcommand{\result}[1]{\textcolor{black}{#1}}

\newcommand\blfootnote[1]{%
  \begingroup
  \renewcommand\thefootnote{}\footnote{#1}%
  \addtocounter{footnote}{-1}%
  \endgroup
}

\usepackage{atbegshi}

\AtBeginShipout{\AtBeginShipoutUpperLeft{%
  \put(\dimexpr\paperwidth/2-5cm\relax,-1.5cm){\makebox[10cm][c]{To appear in proceedings of International Symposium on Experimental Robotics (ISER 2023), Chiang Mai, Thailand}}%
}}

\begin{document}
\mainmatter              
\title{Physics-Based Object 6D-Pose Estimation \\ during Non-Prehensile Manipulation}
\titlerunning{Physics-based Object Pose Estimation}  
%
\author{Zisong Xu \and Rafael Papallas \and Mehmet R. Dogar}
\authorrunning{Zisong Xu et al.} 
%
\tocauthor{Zisong Xu, Rafael Papallas, and Mehmet Dogar}
\institute{University of Leeds, LS2 9JT, UK\\
\blfootnote{This research has received funding from EPSRC under the grant EP/V052659/1. 
    Associated code and data available at: \href{https://github.com/ZisongXu/trackObjectWithPF}{https://github.com/ZisongXu/trackObjectWithPF}
    For the purpose of open access, the authors have applied a Creative Commons Attribution (CC BY) license to any Author Accepted Manuscript version arising.
    }
\email{\{sc19zx, r.papallas, m.r.dogar\}@leeds.ac.uk}
}

\maketitle              


\begin{abstract}
We propose a method to track the 6D pose of an object over time, while the object is under non-prehensile manipulation by a robot. At any given time during the manipulation of the object, we assume access to the robot joint controls and an image from a camera. We use the robot joint controls to perform a physics-based prediction of how the object might be moving. We then combine this prediction with the observation coming from the camera, to estimate the object pose as accurately as possible. We use a particle filtering approach to combine the control information with the visual information. We compare the proposed method with two baselines: (i) using only an image-based pose estimation system at each time-step, and (ii) a particle filter which does not perform the computationally expensive physics predictions, but assumes the object moves with constant velocity. Our results show that making physics-based predictions is worth the computational cost, resulting in more accurate tracking, and estimating object pose even when the object is not clearly visible to the camera.
\end{abstract}

\section{Introduction}\label{sec:intro}
%


 The dominant approach in robotic manipulation applications has been estimating the object pose from RGB(D) camera images \cite{correll2016analysis}. Object pose is estimated using deep learning (e.g. DOPE \cite{tremblay2018deep}) or feature-based \cite{collet2011moped} methods. 

\begin{figure*}[htb]
\centering
  \vspace{-0.2cm}
  \includegraphics[width=1.0\textwidth]{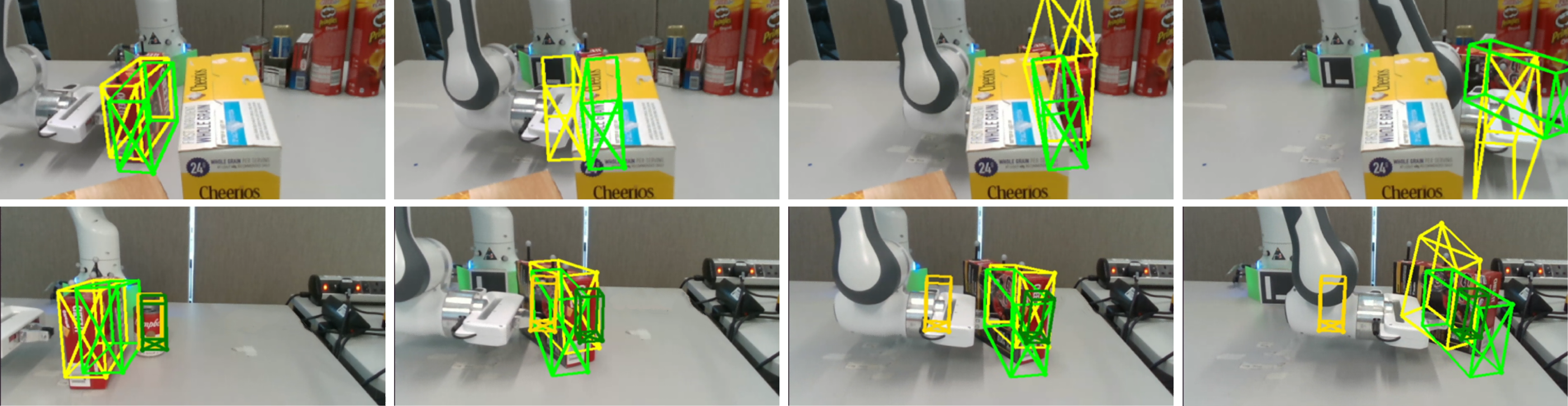}
  \vspace{-0.4cm}
  \caption{Two example scenes. The estimated object pose of our method (PBPF) is shown as a green wireframe box.  The pose from a vision-only system (DOPE) is shown as a yellow wireframe box.
  See the attached video or at: \href{https://youtu.be/B9-5iTwgFnk}{https://youtu.be/B9-5iTwgFnk}.
  } 
  \vspace{-0.4cm}
  \label{s1_obscured}
\end{figure*}

However, during non-prehensile manipulation \cite{ruggiero2018nonprehensile,papallas2020online,papallas2022} in clutter, objects are often fully or partially obstructed from the camera view, by cluttering objects (e.g., Fig.~\ref{s1_obscured}, top-row) or by the robot hand (e.g., Fig.~\ref{s1_obscured}, bottom-row). This degrades the performance of vision-only pose estimation. 

The key insight we have is that, during non-prehensile manipulation, physics-based predictions about objects' motion can be used to supplement the visual information, to better estimate object pose even when there are significant visual occlusions. We propose a particle-filtering method that combines physics-based predictions (using the available robot joint information and a physics-engine) with the visual camera information to estimate the pose of manipulated objects. We show two example scenes in 
Fig.~\ref{s1_obscured}. In the top-row, the robot is pushing a red box. As the pushed object becomes visually obstructed by the box closer to the camera, the vision-only system fails to estimate the object's pose, whereas our method can generate good estimates. In the bottom-row, two objects are manipulated: a red box, which then pushes against a red can. As the push progresses, the robot hand visually obstructs the red box and the red box visually obstructs the can. Our method uses the physics-based interaction between the red box and the can (in addition to the robot-red box interaction) to estimate the poses of both objects.

\PROBLEMFORMULATI{A similar work that combines visual information with physics-based reasoning for 6D pose estimation in clutter is \cite{mitash2019physics}. However, this work addresses static scenes (e.g. based on the physical stability of a pile of objects). Instead, our work uses physics predictions as the robot is manipulating/moving the objects. Other work also uses physics and combines it with contact/tactile sensing \cite{koval2013pose,schmidt2015depth,pfanne2018fusing,zhong2022soft}, as robot-object contact forces are tightly related to the pose and motion of the object. In our work, we do not assume we have contact sensing, but instead, we use the more commonly available camera sensing. }

Other vision-based object tracking systems exist, sometimes using particle filtering similar to our work \cite{pauwels2015simtrack,pauwels2014real,ge2021vipose,lu2022online}, but they either assume the object does not move \cite{choi2013rgb} (but the camera may move as in \cite{wen2020se}), or they \cite{deng2021poserbpf} assume constant velocity. \PROBLEMFORMULATI{Recent work from \cite{jongeneel2022model} proposes a method to track the pose of a box undergoing a non-smooth impact with a surface (e.g. a box tossed onto a surface) using a camera. The work derives and uses a specific impact model for the box-surface system. The main difference of our work to the above is that, in addition to visual information, we use robot control information and its physical effects.} To the best of our knowledge, this is the first work that integrates general physics-based predictions with pose estimation.


\section{Problem Formulation}\label{formulation}


We use $x_t \in SE(3)$ to represent the full 6D pose of the object at time $t$. Our objective is to estimate $x_t$ for all times $t$ during manipulation, i.e. to track the object pose as it is being manipulated. At any time $t$, we assume we have access to:
\begin{itemize}
    \item the controls $u_t$, which are the joint-space controls executed by the robot since the last time-step $t-1$; 
    \item the observation $z_t$, which is an image from a camera looking at the scene.
\end{itemize}

Therefore, our problem can be formalized as: At any time $t$, given all controls since the beginning of the manipulation $\{u_0, u_1, \ldots,u_t\}$ and all observations $\{z_0, z_1, \ldots,z_t\}$, predict an estimate of the object pose, $\tilde{x}_t$.

We use $x^*_t$ to represent the ground truth pose of the object. We define the \textit{positional error} as the Euclidean distance between the positional components of $\tilde{x}_t$ and $x^*_t$. We define the \textit{rotational error} as the (smaller) angle required to represent the relative rotation of $\tilde{x}_t$ and $x^*_t$ around a certain axis: Given $\tilde{q}_t$ and $q^*_t$ as the quaternion components of the estimated and ground truth poses, respectively, the rotational error is given by the amount of rotation suggested by $q^*_t * \tilde{q}^{-1}_t$. Our goal is to minimize positional and rotational errors.

Below, we discuss three different methods (two baselines and our proposed method): (i) Repetitive single-snapshot pose estimation (baseline, in Sec.~\ref{sec:single_snapshot}), (ii) Physics-based particle filtering (proposed method, in Sec.~\ref{sec:physics_based}), and (iii) Constant-velocity particle filtering (baseline, in Sec.~\ref{sec:constant_velocity}).


\section{Repetitive Single-Snapshot Pose Estimation}\label{sec:single_snapshot}

One commonly used method in robotic manipulation is to use a system that can estimate the pose of the object from a single-snapshot, i.e. using only the current camera image $z_t$. There are multiple systems developed to perform object pose estimation given an image, as mentioned in Sec.~\ref{sec:intro}. In this paper, we use DOPE \cite{tremblay2018deep} as a state-of-the-art deep-learning-based pose estimation system. DOPE has been trained on the YCB objects \cite{calli2015benchmarking} that we also use, and is publicly available. Therefore, the repetitive single-snapshot pose estimation method, in our case, corresponds to running DOPE at every time-step $t$ during manipulation: $\tilde{x}_t = DOPE(z_t)$. This method treats every new observation as independent from the previous time steps and therefore does not use temporal continuity. The performance particularly degrades when the object is partially or fully obstructed in the camera view. Moreover, it does not use the control information.


\section{Physics-Based Particle Filtering (PBPF)}\label{sec:physics_based}


Using a Bayesian approach, we estimate the probability distribution for $x_t$ given all the previous controls and observations, $p(x_t \, | \, z_0,z_1,...,z_t,u_0,u_1,...,u_t)$, i.e., the \textit{belief state}. In particle filtering \cite{thrun2006probabilistic}, the belief state is represented using a set of \textit{particles} sampled from this distribution,
$\mathcal{X}_{t} := x_{t}^{[1]}, x_{t}^{[2]}, ..., x_{t}^{[M]}$, 
where each particle $x_{t}^{[m]} \, (1 \leq m \leq M)$ is an instance of the state at time $t$. In our setting, each particle represents a possible pose for all the manipulated objects. 

At any time $t$, if a single estimate $\tilde{x}_t$ is required (e.g. the estimates shown in Fig.~\ref{s1_obscured}), we use the mean of all the particles in $\mathcal{X}_t$ to compute $\tilde{x}_t$. 

During particle filtering, at each time-step $t$, the previous set of particles $\mathcal{X}_{t-1}$ are updated using the current controls, $u_t$, and observation, $z_t$, to generate a new set of particles $\mathcal{X}_{t}$. This happens in two stages: the \textit{motion update} (presented in Sec.~\ref{sec:motion_update}), and the \textit{observation update} (in Sec.~\ref{sec:observation_update}).

\subsection{Motion Update}\label{sec:motion_update}

During this first stage, for each particle $x_{t-1}^{[m]}$, we generate a new intermediate particle (shown as $\Bar{x}_t^{[m]}$), by sampling $\Bar{x}_{t}^{[m]} \sim p(x_t \, | \, x_{t-1}^{[m]},u_{t})$.

We assume access to a physics engine, $f_{\theta}$, such that:
$x_{t} = f_{\theta}(x_{t-1},u_{t})$. Here, $\theta$ refers to physical parameters, e.g. friction coefficient at the contacts, restitution of the contacts, and the mass of the object. The physics engine is deterministic, and therefore cannot directly be used instead of the probabilistic motion model. Therefore,  we approximate sampling from the motion model $\Bar{x}^{[m]}_{t} \sim p(x_t \, | \, x_{t-1}^{[m]},u_{t})$, by first sampling $\theta$ from a distribution representing our uncertainty about the physical parameters of the objects, $\theta_t^{[m]} \sim \mathcal{N}(\mu_{\theta},\sigma_{\theta}^2)$,
and then running the physics engine with the sampled parameter:    $\Bar{x}^{[m]}_{t} = f_{\theta_t^{[m]}}(x_{t-1}^{[m]},u_{t}) + \epsilon$, 
with the addition of small Gaussian noise $\epsilon \sim \mathcal{N}(0,\sigma^2_{f})$.

\subsection{Observation Update}\label{sec:observation_update}

During this second stage of particle filtering, for each intermediate particle $\Bar{x}^{[m]}_{t}$, we calculate an \textit{importance factor} $w_t^{[m]}$ using the observation, $w_{t}^{[m]} = p(z_t \, | \, \Bar{x}_{t}^{[m]})$. After the importance factors for all the particles are computed, they are used to \textit{re-sample} the new set of particles $\mathcal{X}_t$, completing the particle filter update.

To compute $ p(z_t \, | \, \Bar{x}_{t}^{[m]})$, we propose to use the vision-only pose estimation system (e.g., DOPE) to predict the pose of the object according to $z_t$, and then to use the distance of $\Bar{x}_{t}^{[m]}$ to this predicted pose to compute a probability value, $ p(z_t \, | \, \Bar{x}_{t}^{[m]}) \simeq \mathcal{N}(\Bar{x}_{t}^{[m]}; \, DOPE(z_t), \sigma^2_{DOPE})$, where $DOPE(z_t)$ is the pose predicted by DOPE, and the parameter $\sigma^2_{DOPE}$ represents the variance of DOPE errors for the object. This can be estimated beforehand for an object by collecting DOPE estimates for the object and comparing it to a ground truth pose.

\subsection{Using Visibility Information}\label{sec:visibility}

The observation model above assumes that a DOPE detection, $DOPE(z_t)$, is always available. However, for a given camera image $z_t$, DOPE may not output a pose estimate. At such time points, one option is to skip the observation update. This is what we do in the basic PBPF algorithm.

However, the lack of detection of the object is not a lack of information. The object is probably partially or fully occluded. Therefore, we also implemented an extension of our algorithm:  \textit{PBPF with Visibility} (PBPF-V), which is currently implemented only for single-object manipulation.

For this method, we first compute a \textit{visibility score}, $v_t^{[m]}$, for each particle. To compute the visibility score, in the physics engine, we shoot $R$ rays from the camera towards $R$ points uniformly sampled on the manipulated object. Some of these rays hit the object, but some others may hit other occluding objects or robot links before hitting the manipulated object. We define the visibility of a particle as the ratio of rays hitting the particle to the number of all rays:
$	v_t^{[m]} =  \frac{\text{number of rays hitting the particle}}{R}
$.

A visibility score of $1$ implies perfect visibility of the object by the camera, whereas a score of $0$ implies the particle is completely occluded.  Then, we can relate the importance factor of a particle, ${w}_{t}^{[m]}$, to its visibility $v_t^{[m]}$ based on whether a DOPE estimate is available or not at time $t$. The particular method to relate these values depends on a model of DOPE behaviour under occlusion. Our experiments with DOPE did \textbf{not} suggest that DOPE gradually becomes worse with more occlusion. Instead, after a certain amount of occlusion, DOPE usually loses the object completely. Therefore, instead of relating ${w}_{t}^{[m]}$ and $v_t^{[m]}$ (inversely) proportionally, we implemented a threshold-based model: If DOPE is not able to detect the object at time $t$, then we assign a fixed high importance factor value ($W_H$) to particles that have low visibility scores (i.e. $v_t^{[m]} < V_H$, where $V_H$ is a fixed visibility threshold), and a fixed low importance factor value ($W_L$) to particles that have high visibility scores (i.e. $v_t^{[m]} > V_H$). Conversely, if a DOPE estimate is available at time $t$, then we compute the importance factor values according to the previous Section~\ref{sec:observation_update}, but penalize the particles with low visibility ($v_t^{[m]} < V_L$, where $V_L$ is a fixed low visibility threshold) by scaling their importance factor by $\alpha_W \in [0,1]$. 


\section{Constant-Velocity Particle Filtering (CVPF)}\label{sec:constant_velocity}

Our first baseline method was presented in Sec.~\ref{sec:single_snapshot}, which estimates the object’s pose from a new image frame at every timestep. Our second baseline method, presented here, is a particle filter, similar to PBPF presented in Sec.~\ref{sec:physics_based}, the only difference being in the \textit{motion update} stage. While PBPF uses a physics engine in its motion model, here, instead we use a computationally cheap motion model, which simply assumes that the object moves with a constant velocity. In other words, during the motion update, we update a particle at time step $t$ using the estimated pose of the object from the previous time steps $t-1$ and $t-2$.
We calculate the ``difference'' between the estimated pose of the object at the previous time-steps: $\mathrm{d}x = \tilde{x}_{t-1} - \tilde{x}_{t-2}$
and then, we assume the object keeps moving with the same velocity, and use the motion update rule: $\Bar{x}_{t}^{[m]} = {x}_{t-1}^{[m]} + \mathrm{d}x + \epsilon$
where $\epsilon$ is an extra noise term, similar to the parameter in \ref{sec:motion_update}. 

There are no differences between PBPF and CVPF handling the pose information from DOPE.

\section{Experiments And Results}

We evaluated and compared the performance of our methods in eight different non-prehensile manipulation scenes. (All scenes can be seen in the attached video.)
\dope{We used three objects from the YCB dataset \cite{calli2015benchmarking}, the CheezIt box, the Gelatin Jell-o and the Campbell Soup, a cylinder. DOPE had good performance of these objects when they were visible.}

\begin{figure*}[!htb]
\centering
  \includegraphics[width=1.0\textwidth]{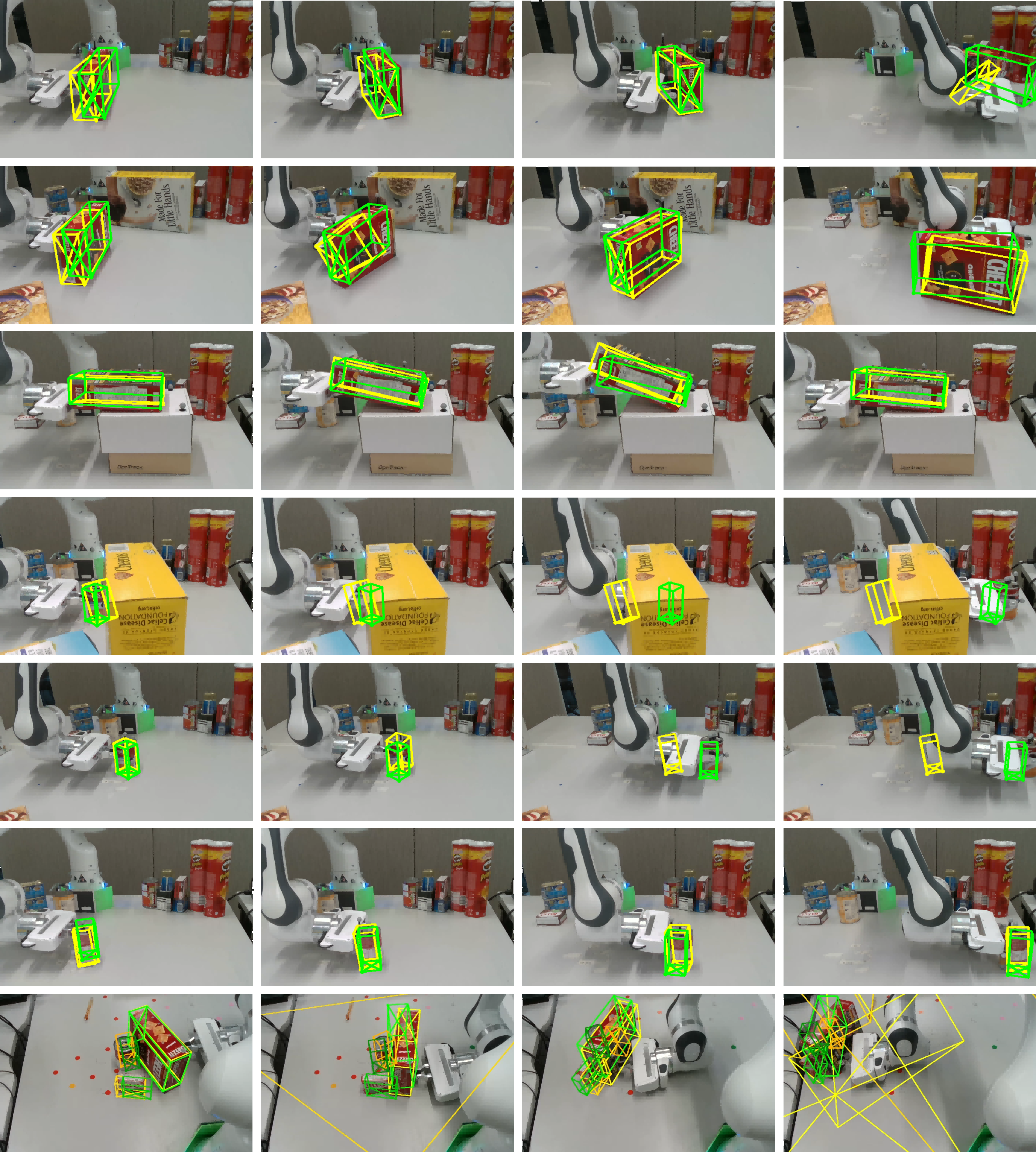}
  \caption{Experimental Scenes. The images are from the tracking camera. The estimated object pose of our method (PBPF) is shown as a green wireframe box, while the estimated object pose from DOPE is shown as a yellow wireframe box.} 
  \label{real_image_allall}
\end{figure*}

\cvpf{\noindent\textbf{Cheezit~Scene~1} (shown in Fig.~\ref{s1_obscured}-first row) and \textbf{Soup~Scene~1} (shown in Fig.~\ref{real_image_allall}-fourth row): The robot pushes the object among the clutter, where the cluttering object obstructs the view of the camera. The pose of the obstructing object is fixed and known.}

\cvpf{\noindent\textbf{Cheezit~Scene~2} (shown in Fig.~\ref{real_image_allall}-first row) and \textbf{Soup~Scene~2} (shown in Fig.~\ref{real_image_allall}-fifth row): The robot pushes the object on a clear table. The hand can obstruct the camera view during pushing.}

\cvpf{\noindent\textbf{Cheezit~Scene~3} (shown in Fig.~\ref{real_image_allall}-second row) and \textbf{Soup~Scene~3} (shown in Fig.~\ref{real_image_allall}-sixth row): The robot pushes the object on a clear table, and the camera has a clear view of the object.}

\cvpf{\noindent\textbf{Cheezit~Scene~4} (shown in Fig.~\ref{real_image_allall}-third row): The robot tilts the object up and then brings it down. The camera has a clear view. The pose of the supporting object is fixed and known.}

\cvpf{\noindent\textbf{Scene~5} (shown in Fig.~\ref{real_image_allall}-seventh row): The robot pushes multiple objects. These objects occlude each other during the push. We track all objects' poses simultaneously.}

In Scenes 1, 2, and 3, obstacles in the clutter do not physically interact with the robot and the tracked object.

We have implemented our methods as below\footnote{Experiments were performed on CPU: 11th Gen Intel(R) Core(TM) i9-11900@2.50GHz; GPU: NVIDIA GeForce RTX 3090; RAM: 128580 Mb}.


\textbf{PBPF}: The physics-based particle filtering method (Sec.\ref{sec:physics_based}). (a) \textit{Motion model parameters.} $\mu_{\theta}$ and $\sigma_{\theta}$: Mean friction coefficient of 0.1 and standard deviation of 0.3, with minimum capped at 0.001. Mean restitution of 0.9 and standard deviation of 0.2. The mean mass of 0.38 kg and the standard deviation of 0.5 with a minimum cap of 0.05 kg. $\sigma_{f}$: For position 0.005 m, for rotation 0.05 rad. As the physics model, $f_\theta$, we used the Pybullet physics engine \cite{coumans2019}. A Pybullet environment was initialized for each particle. The Pybullet environments for particles were parallelized over the 8 (16 virtual) CPU cores (11th Gen Intel(R) Core(TM) i9-11900@2.50GHz) of the computer. (b) \textit{Observation model parameters.} $\sigma_{DOPE}$: For position 0.02 m and rotation 0.09 rad. (c) \textit{Update frequency}. $\Delta t = 0.16 s$. (d) \dope{\textit{Number of particles.} $M=70$.} (e) \textit{Initialization}. We use a Gaussian distribution to initialize 70 particles at $t=0$. The mean pose is estimated by DOPE at $t=0$. The standard dev. for initialization is 0.16 m and 0.43 rad.

\textbf{PBPF-V}: An extension of PBPF which also considers visibility  (as described in Sec.~\ref{sec:visibility}). \dope{All parameters are the same with PBPF except that the number of particles $M=150$ in Scene 5}. Additional visibility parameters: \dope{For Cheezit and Gelatin}, $W_H=0.75$, $W_L=0.25$, $V_H=0.95$, $V_L=0.5$, $\alpha_W=0.33$. For Soup, $W_H=0.6$, $W_L=0.45$, $V_H=0.8$, $V_L=0.9$, $\alpha_W=0.33$.

\textbf{CVPF}: The constant-velocity particle filtering method. All parameters were the same with PBPF, except the number of particles, $M$. Since CVPF is computationally cheaper, it could handle more particles $M=150$.

\textbf{Ground truth}: We used a marker-based OptiTrack
system to record ``ground truth'' poses, $x^{*}_t$, during manipulation.

\begin{figure}[htb]
\centering
  \includegraphics[width=1.0\textwidth]{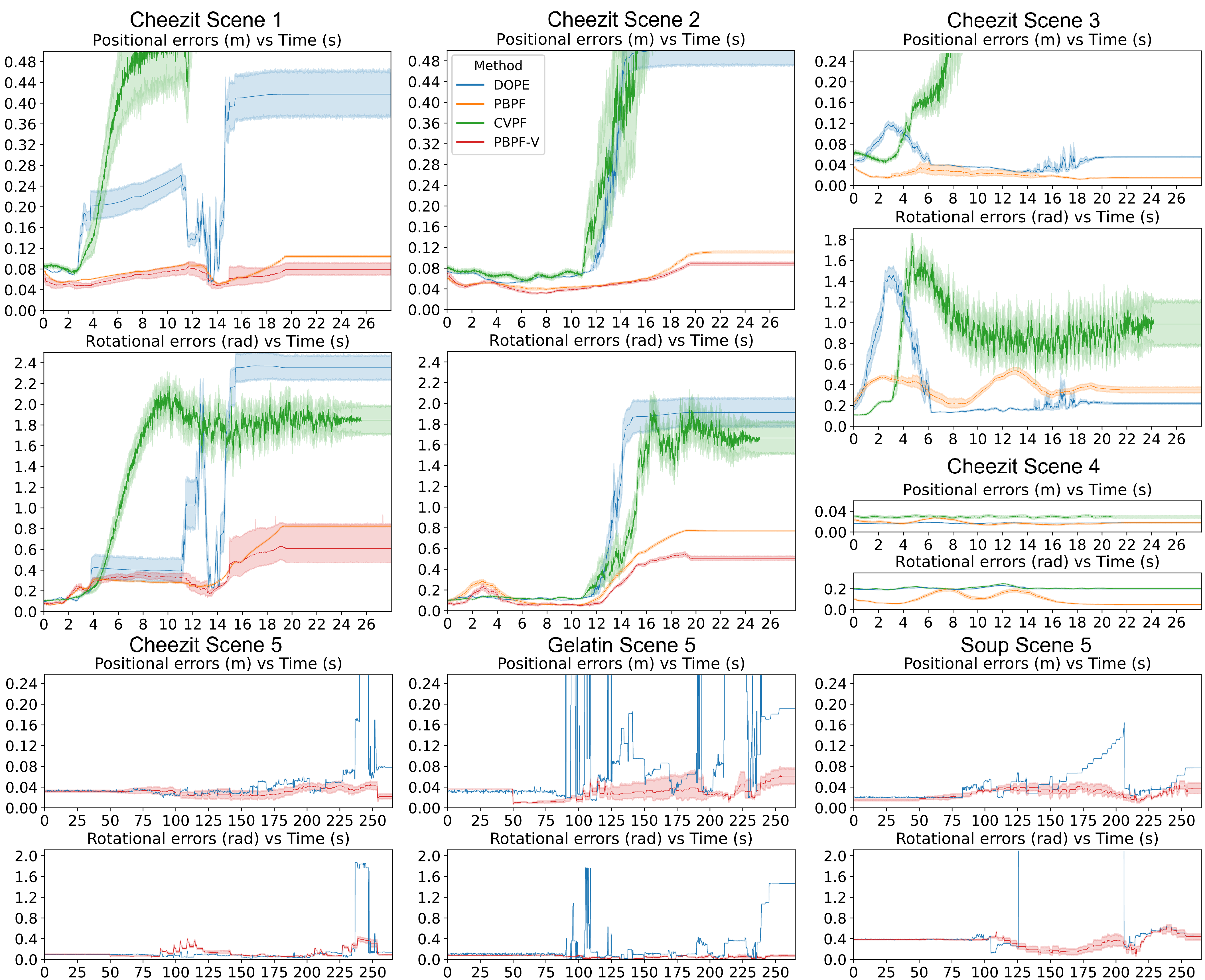}
  \caption{\capture{Positional and rotational errors of different methods in four different scenes with the Cheezit object and the fifth scene shows the tracking of three different target objects, each with its own positional and rotational errors in different methods. Plots show mean values with the shadows indicating the 95\% Confidence Interval. In all plots, the horizontal axes show time, and the vertical axes show the positional/rotational error. Similar plots for all scenes and all objects can be seen in our repository. 
  }}
  \label{results_figure}
\end{figure}

\begin{table*}
\fontsize{11pt}{11pt}\selectfont
\renewcommand\arraystretch{1}
\caption{Mean and standard deviation of the errors of different methods}\label{results_table_single}
\resizebox{\textwidth}{17mm}{
\begin{tabular*}{\textheight}{@{\extracolsep\fill}lcccccccccccccccc}
& \multicolumn{4}{@{}c@{}}{\textbf{Scene 1}}& \multicolumn{4}{@{}c@{}}{\textbf{Scene 2}}& \multicolumn{4}{@{}c@{}}{\textbf{Scene 3}}& \multicolumn{4}{@{}c@{}}{\textbf{Scene 4}} \\
& \multicolumn{2}{@{}c@{}}{pos.err(m)}& \multicolumn{2}{@{}c@{}}{rot.err(rad)}& \multicolumn{2}{@{}c@{}}{pos.err(m)}& \multicolumn{2}{@{}c@{}}{rot.err(rad)}& \multicolumn{2}{@{}c@{}}{pos.err(m)}& \multicolumn{2}{@{}c@{}}{rot.err(rad)}& \multicolumn{2}{@{}c@{}}{pos.err(m)}& \multicolumn{2}{@{}c@{}}{rot.err(rad)} \\
& $\mu$	& $\sigma$ & $\mu$	& $\sigma$& $\mu$	& $\sigma$& $\mu$	& $\sigma$& $\mu$	& $\sigma$& $\mu$	& $\sigma$& $\mu$	& $\sigma$& $\mu$	& $\sigma$ \\
\midrule 

\textbf{Cheezit} & & & & & & & & & & & & & & & & \\

PBPF   & 0.081 & 0.020 & 0.484 & 0.272 & 0.070 & 0.030 & 0.426 & 0.309 & \textbf{0.019} & 0.024 & \textbf{0.367} & 0.187 & 0.018 & 0.008 & \textbf{0.092} & 0.067 \\
 
PBPF-V & \textbf{0.063} & 0.020 & \textbf{0.396} & 0.295 & \textbf{0.058} & 0.021 & \textbf{0.294} & 0.204 &   -   &   -   &   -   &   -   &   -   &   -   &   -   &   -   \\

DOPE   & 0.289 & 0.218 & 1.345 & 1.168 & 0.293 & 0.248 & 1.046 & 1.027 & 0.052 & 0.033 & 0.375 & 0.447 & \textbf{0.017} & 0.002 & 0.201 & 0.013 \\

CVPF   & 2.359 & 7.421 & 1.472 & 0.951 & 0.407 & 0.678 & 0.878 & 0.952 & 25.14 & 152.8 & 0.879 & 0.952 & 0.029 & 0.012 & 0.207 & 0.022 \\

\midrule

\textbf{Soup} & & & & & & & & & & & & & & & & \\

PBPF   & \textbf{0.043} & 0.023 & 1.133 & 0.364 & 0.049 & 0.032 & 0.851 & 0.249 & \textbf{0.026} & 0.021 & \textbf{0.405} & 0.455 &   -   &   -   &   -   &   -   \\

PBPF-V & 0.045 & 0.027 & \textbf{1.077} & 0.395 & \textbf{0.046} & 0.028 & \textbf{0.724} & 0.262 &   -   &   -   &   -   &   -   &   -   &   -   &   -   &   -   \\

DOPE   & 0.215 & 0.117 & 1.174 & 0.372 & 0.206 & 0.112 & 0.776 & 0.222 & 0.071 & 0.140 & 0.438 & 0.482 &   -   &   -   &   -   &   -   \\

CVPF   & 0.298 & 0.153 & 1.884 & 0.794 & 0.240 & 0.126 & 1.702 & 0.845 & 6.528 & 52.09 & 1.174 & 0.982 &   -   &   -   &   -   &   -   \\
\midrule
\end{tabular*}}
\end{table*}

\subsection{Experimental Procedure}


\dope{In Scenes 1-4, we repeated 10 robot runs, and in Scene 5, we repeated 1 robot run (i.e. a total of 71 real robot executions)}. During each run, we recorded the robot controls, $u_t$, and the camera images, $z_t$. Since particle filtering is a sampling-based method, it can generate different results with the same input. \dope{For statistical accuracy, we evaluated each method 10 times on the data from each of the 10 runs in Scenes 1-4, giving 100 evaluations of each method for Scenes 1-4. In Scene 5, we evaluated each method 20 times on the data from the 1 real robot execution.} 


At each time step of each evaluation, we computed the positional and rotational errors of the mean estimate, against the ground truth. When computing errors for DOPE, if DOPE did not output any pose at a certain time step (e.g., because of occlusions), we used the latest pose reported before that time step.

\subsection{Results}

\dope{We present the overall (averaged over all runs and all time-steps) positional and rotational errors from Scene 1-4 in Table.~\ref{results_table_single}, and from Scene 5 in Table.~\ref{results_table2_multi}}. Compared to DOPE and CVPF, PBPF performs significantly better in Scenes 1, 2, and 3. In Scene 4, where DOPE has a good view and overall good detection, PBPF and DOPE perform similarly. CVPF performs worse out of all the methods. \dope{In Scene 5, we only ran the PBPF-V algorithm used for comparison with DOPE. PBPF-V performs significantly better than the baseline method.}

\begin{table*}[!b]
\fontsize{11pt}{11pt}\selectfont
\renewcommand\arraystretch{1}
\caption{Mean and standard deviation of the  errors when tracking multi-object (Scene 5)}\label{results_table2_multi}
\resizebox{\textwidth}{9mm}{
\begin{tabular*}{\textheight}{@{\extracolsep\fill}lcccccccccccc}
& \multicolumn{4}{@{}c@{}}{\textbf{Cheezit}}& \multicolumn{4}{@{}c@{}}{\textbf{Gelatin}}& \multicolumn{4}{@{}c@{}}{\textbf{Soup}} \\
& \multicolumn{2}{@{}c@{}}{pos.err(m)}& \multicolumn{2}{@{}c@{}}{rot.err(rad)}& \multicolumn{2}{@{}c@{}}{pos.err(m)}& \multicolumn{2}{@{}c@{}}{rot.err(rad)}& \multicolumn{2}{@{}c@{}}{pos.err(m)}& \multicolumn{2}{@{}c@{}}{rot.err(rad)} \\
& $\mu$	& $\sigma$ & $\mu$	& $\sigma$& $\mu$	& $\sigma$& $\mu$	& $\sigma$& $\mu$	& $\sigma$& $\mu$	& $\sigma$\\
\midrule 
PBPF-V & \textbf{0.031} & 0.011 & \textbf{0.126} & 0.087 & \textbf{0.031} & 0.023 & \textbf{0.056} & 0.030 & \textbf{0.027} & 0.013 &   \textbf{0.346} & 0.139 \\
DOPE   & 0.058 & 0.108 & 0.158 & 0.337                   & 0.179 & 0.665 & 0.272 & 0.410                   & 0.045 & 0.029 & 1.193 & 1.192 \\
\midrule
\end{tabular*}}
\end{table*}

When there is occlusion (Scenes 1 and 2), PBPF-V performs better than basic PBPF for the Cheezit object. For the Soup object, the difference between PBPF and PBPF-V is not significant. This is probably due to the Soup being a much smaller object, for which DOPE's behaviour under occlusion is much more difficult to model accurately.

We also present the positional \result{and rotational }errors of each method over the course of manipulation in Scenes 1-4 for Cheezit in Fig.~\ref{results_figure}. (Results for Soup are similar, and are in our repository.) We can see that PBPF and PBPF-V produce better and more stable estimates of the object position. In Fig.~\ref{results_figure}, we also show the results for each object in Scene 5. PBPF-V significantly outperforms DOPE.

\result{It is worth noting that even with a good field of view, DOPE can be noisy, because it uses a single visual snapshot, which becomes unstable when the light, angle, and other factors change. This can lead to physically unrealistic poses, e.g. target object can be inside the robot arm, or two target objects can penetrate. In contrast, our method (PBPF) outputs physically feasible poses,  effectively avoiding the unrealistic scenarios above.}

\end{document}